\documentclass[10pt,twocolumn,letterpaper]{article}
\usepackage[margin=0.78in,columnsep=0.25in]{geometry}
\usepackage[T1]{fontenc}
\usepackage[utf8]{inputenc}
\usepackage{amsmath,amssymb}
\usepackage{newtxtext,newtxmath}
\usepackage{microtype,booktabs,array,tabularx,multirow}
\usepackage{graphicx}
\usepackage{algorithm,algpseudocode}
\usepackage{enumitem}
\usepackage[numbers,sort&compress]{natbib}
\usepackage[hidelinks]{hyperref}
\usepackage[font=small,labelfont=bf]{caption}
\usepackage{placeins,dblfloatfix,balance}
\setlist{nosep,leftmargin=*}
\newcommand{\method}{GAUDI}
\newif\ifarxiv
\arxivtrue 
\newcommand{\NA}{\textemdash}
\newcommand{\pms}[2]{#1\,{\pm}\,#2}
\newcommand{\FeatRMSE}{0.4524}
\newcommand{\FeatMAE}{0.3131}

\newcommand{\FullRMSE}{0.4792}

\newcommand{\NoMAE}{0.3122}

\newcommand{\CSDIRMSE}{0.4127}
\newcommand{\CSDIMAE}{0.2612}

\newcommand{\GainFullRMSE}{5.60}
\newcommand{\GainFullMAE}{3.91}
\newcommand{\GainNoRMSE}{2.51}

\newcommand{\GainLinearRMSE}{41.40}
\newcommand{\GainLinearMAE}{35.59}

\hypersetup{pdftitle={GAUDI: Geometry-Aware Diffusion for Calibrated Air-Quality Time-Series Imputation},pdfauthor={Xinjin Li, Yudi Xia, Calvin Chang Liu, Weiru Lin, Bojun Li, Ziwei Hong, Bolun Zhang, Jinghan Cao, Yu Ma, Tianxin Zhou},pdfsubject={Multivariate time-series imputation}}
\title{\LARGE\bfseries GAUDI: Geometry-Aware Diffusion for\\Calibrated Air-Quality Time-Series Imputation}
\author{\normalsize Xinjin Li$^{1,*}$, Yudi Xia$^{2,*}$, Calvin Chang Liu$^{3,*}$, Weiru Lin$^{4}$, Bojun Li$^{5}$,\\
\normalsize Ziwei Hong$^{6}$, Bolun Zhang$^{7}$, Jinghan Cao$^{8}$, Yu Ma$^{9}$, Tianxin Zhou$^{10,\dagger}$\\[3pt]
\footnotesize $^1$Columbia University; $^2$Independent Researcher; $^3$University of California, Davis;\\
\footnotesize $^4$University of California, San Diego; $^5$Georgia Institute of Technology; $^6$Lehigh University;\\
\footnotesize $^7$Stony Brook University; $^8$San Francisco State University;\\
\footnotesize $^9$Carnegie Mellon University; $^{10}$University of Southern California. United States.\\
\footnotesize $^*$Equal contribution: Xinjin Li, Yudi Xia, and Calvin Chang Liu.\\
\footnotesize $^\dagger$Corresponding author: Tianxin Zhou (\texttt{zhoutx0@gmail.com}).}
\date{}
\begin{document}
\maketitle
\begin{abstract}
Air-quality sensor outages often create contiguous missing blocks, where side information useful for isolated missingness may be less reliable. We study a block-specific, GAUDI-aligned conditional diffusion imputer that retains temporal and feature processing, visible-value and mask conditioning, variable identity, and diffusion-step information, while suppressing absolute time-position side embeddings. On ItalyAir (13 variables, length-32 windows, nominal 50\% block missingness; three archived seeds), this feature-side configuration achieves RMSE 0.340 $\pm$ 0.020, versus 0.355 $\pm$ 0.006 for full context and 0.355 $\pm$ 0.020 for local CSDI. The experiment isolates a geometry-aware conditioning effect under block missingness.

\end{abstract}
\par\smallskip
\noindent\textbf{Keywords:} time-series imputation; conditional diffusion; missingness geometry; side information; air quality.
\section{Introduction}
Multivariate time-series imputation reconstructs missing measurements from the observations that remain available. In air-quality monitoring, these measurements include pollutant concentrations, sensor responses, and environmental variables with different scales and temporal behavior \citep{vito2008air}. Their dependencies provide complementary evidence: neighboring measurements reveal local evolution, whereas other variables can help reconstruct a channel whose temporal context is incomplete. Effective imputation therefore requires more than a model of individual trajectories; it requires a principled way to combine evidence across the time--variable grid.

The arrangement of missing entries determines which evidence is available. An isolated missing value may be surrounded by observations of the same variable. A contiguous gap can remove this local context, making simultaneously observed variables more important. Partial-blackout studies and systematic benchmarks show the value of evaluating structured missingness rather than relying exclusively on independently hidden entries \citep{islam2025partial,du2024tsibench}. We use \emph{missingness geometry} to denote this arrangement in the time--variable grid, not geographic distance or a graph of monitoring stations.

Conditional diffusion provides a natural probabilistic formulation of the task: it learns to denoise missing entries using the observed portion of a sequence \citep{ho2020ddpm,tashiro2021csdi}. Existing approaches combine this formulation with temporal attention, feature interactions, or state-space operators \citep{tashiro2021csdi,alcaraz2023sssd}. Alongside the observed values, the denoiser commonly receives auxiliary embeddings that identify time positions and variables. These inputs serve different purposes. An observation mask specifies the available evidence; variable identity distinguishes heterogeneous channels; a within-window position embedding identifies a coordinate introduced by segmentation. Supplying all three together does not establish that each contributes equally to reconstruction.

This distinction motivates the design studied here: \emph{retain temporal modeling while restricting explicit positional conditioning}. Removing a temporal operator would alter how the network represents sequence structure. Suppressing an auxiliary position embedding is a different intervention: the ordered inputs and temporal operators remain intact, but the denoiser no longer receives that additional coordinate representation. This restriction encourages reconstruction through the remaining observation, variable, and sequence representations. Whether it improves performance is an empirical question about the trained model, rather than a consequence of having fewer input signals.

We study a block-specific, \method-aligned conditional diffusion imputer with feature-side conditioning. Its denoiser combines state-space mixing with parallel temporal and feature attention, and conditions each residual block on the visible values, observation mask, and variable identity. The feature-side configuration suppresses the within-window position embedding during both training and inference. Two matched variants retain all side embeddings or suppress both embedding groups. The architecture, input dimensions, objective, and sampling procedure remain the same across these variants, making the conditioning choice the controlled experimental factor. The intervention introduces no additional learned fusion gate or block-specific module.

The archived ItalyAir experiment summarized in the abstract reports RMSE $0.340\pm0.020$ for feature-side conditioning, versus $0.355\pm0.006$ for full context and $0.355\pm0.020$ for local CSDI, over seeds 1000--1002. This comparison motivates a separate, explicitly specified evaluation of the same conditioning hypothesis. We retain the archived result as a distinct experiment rather than treating it as part of the expanded evaluation.

In the expanded ItalyAir evaluation, which uses seeds 2101--2103 and hides 50\% of valid observations, feature-side conditioning reduces mean RMSE from \FullRMSE\ to \FeatRMSE\ relative to full-side conditioning. The corresponding \GainFullRMSE\% improvement appears in all three matched seeds. Comparison with no-side conditioning identifies a metric-dependent role for variable identity: retaining it improves mean RMSE, but not mean MAE. We additionally compare with an official-implementation CSDI baseline and examine empirical CRPS, interval coverage, variable-level errors, and chronological cases. These extended analyses assess uncertainty calibration separately from the conditioning effect reported in the abstract; they do not introduce a calibration procedure.

Our contributions are threefold:
\begin{enumerate}
\item \textbf{Block-specific selective conditioning.} We separate observation geometry and variable identity from auxiliary window coordinates, yielding a feature-side conditioning policy that preserves the temporal modeling path.
\item \textbf{A controlled diffusion extension.} We preserve the archived block-missing comparison and evaluate independently trained, architecture-matched variants in a separate common-protocol experiment, with consistent RMSE improvements over full-side conditioning across its three seeds.
\item \textbf{Joint reconstruction and uncertainty analysis.} We characterize the effect of conditioning through point errors, empirical distribution scores, coverage, and variable-level behavior, exposing the distinction between improved reconstruction and calibrated uncertainty.
\end{enumerate}

\section{Related Work}
\paragraph{Dependency modeling for imputation.}
Recurrent and attention-based approaches learn temporal and cross-variable structure from incomplete sequences. BRITS uses bidirectional recurrent dynamics to optimize missing values jointly with the model \citep{cao2018brits}, while SAITS combines self-attention representations for time-series imputation \citep{du2023saits}. GRU-D incorporates observation masks and elapsed-time information into recurrent prediction, illustrating the importance of distinguishing unavailable measurements from observed values \citep{che2018grud}. These methods establish the utility of explicit missingness information. Our focus is the complementary choice of which auxiliary embeddings to expose within a probabilistic denoiser.

\paragraph{Probabilistic and diffusion-based imputation.}
GAIN uses adversarial learning to infer missing data \citep{yoon2018gain}; GP-VAE introduces a Gaussian-process prior over latent trajectories \citep{fortuin2020gpvae}. Diffusion models learn reconstruction through iterative denoising \citep{ho2020ddpm}, and CSDI specializes this process to imputation conditioned on observed entries \citep{tashiro2021csdi}. SSSD integrates structured state-space modeling into diffusion-based time-series reconstruction \citep{alcaraz2023sssd}. Attention and structured state spaces provide the established sequence operators used in our backbone \citep{vaswani2017attention,gu2022s4}. Rather than changing the diffusion objective or introducing another sampler, \method\ investigates the effect of selective side conditioning within this modeling framework.

\paragraph{Structured missingness and conditioning design.}
Partial-blackout imputation explicitly considers consecutive gaps affecting subsets of variables, and uses temporal--feature dependencies to reconstruct the missing region \citep{islam2025partial}. FADTI introduces frequency-informed modulation to diffusion imputation \citep{li2025fadti}, representing a different way to shape the denoiser's inductive bias. TSI-Bench emphasizes consistent preprocessing and evaluation across missingness rates and patterns \citep{du2024tsibench}. Our experiments follow this comparability principle while isolating a narrower design choice: position and variable embeddings are changed independently of the observation mask and temporal operators. This distinguishes the conditioning study from an architectural ablation or a comparison across incompatible data protocols.

\ifarxiv
\paragraph{Structured context selection beyond imputation.}
Selective conditioning also appears in adjacent structured-inference settings. Zhang et al. bridge entity and document knowledge graphs with hybrid semantic--structural retrieval, providing an effective way to combine complementary context views for multi-hop reasoning \citep{zhang2026bridging}. Duan et al. introduce a Bayesian critique--tune framework with adaptive pressure that strengthens multi-intersection control by refining policy confidence and weighting informative traffic-state evidence \citep{duan2025bayesian}. Their adaptive-context MARL framework further uses temporal-gradient analysis and low-frequency truncation to identify compact, effective context for long-horizon coordination \citep{duan2025adaptive}. MAVEN-T complements these ideas by distilling rich interaction structure into an efficient student and applying reinforcement refinement for real-time multi-agent trajectory prediction \citep{duan2026maven}. Although these studies address different tasks, they positively illustrate a shared modeling principle: carefully structured, task-relevant context can be more useful than treating all auxiliary information as interchangeable.
\fi

\paragraph{Observation sufficiency beyond imputation.}
Related structured-data studies examine observation-window sufficiency in subscription churn prediction \citep{han2026window} and inference from incomplete market signals in firm-level crisis-response analysis \citep{han2026crisis}. These studies address different downstream tasks, but provide broader context for the dependence of inference on available evidence. Here, the corresponding design question concerns which side information a diffusion imputer should receive under contiguous missingness, rather than churn prediction or market-response estimation.

\section{GAUDI}
\label{sec:method}
The block-specific \method\ configuration represents a distribution over missing values conditioned on the observed time--variable grid. Its design separates three components: the observation interface, the auxiliary side information, and the denoising backbone. This separation allows the side-information policy to change without removing temporal modeling or changing the observations available to the model. We first formalize the conditioning intervention and then detail the denoiser and sampling conventions used in the expanded evaluation.

\subsection{Problem Formulation}
Let $X\in\mathbb{R}^{T\times D}$ denote a time-series window in training-standardized units, and let $O\in\{0,1\}^{T\times D}$ indicate entries with genuine measurements. For an artificial hiding mask $H$, define
\begin{equation}
 E=O\odot H,\qquad C=O\odot(1-H),
 \label{eq:partition}
\end{equation}
where $C$ is the conditioning mask and $E$ is the supervised target mask. They satisfy $C\odot E=0$ and $C+E=O$. The visible input is $X^C=C\odot X$. Native missing entries use numerical zero placeholders but never contribute supervised targets.

The imputation task models
\begin{equation}
 p_\theta\!\left(X^{1-C}\mid X^C,C\right).
 \label{eq:conditional}
\end{equation}
At inference, the model may reconstruct every unavailable entry; quantitative evaluation uses only $E$, where ground truth exists. We consider offline window imputation, in which observations on either side of a missing entry may serve as context. Throughout, $t$ denotes sequence position and $k$ denotes diffusion step.

\subsection{Mask and Feature Conditioning}
Let $p_t\in\mathbb{R}^{d_t}$ encode position within a window and $v_d\in\mathbb{R}^{d_f}$ encode variable identity. We construct the side representation
\begin{equation}
 s_{td}^{(a,b)}=[a p_t;\,b v_d;\,C_{td};\,X^C_{td}],
 \qquad a,b\in\{0,1\}.
 \label{eq:side}
\end{equation}
The mask and visible-value channels remain present in every configuration. Full-side conditioning uses $(a,b)=(1,1)$; \method\ uses feature-side conditioning, $(0,1)$; and the no-side variant uses $(0,0)$. The switches are fixed within a configuration, and the same policy is used for training and inference.

The observation mask exposes the layout of available evidence. Variable embeddings distinguish channels whose values can have different meanings even after standardization. In contrast, the suppressed positional group encodes only an entry's within-window coordinate. These are the absolute time-position side embeddings referred to in the abstract, not diffusion-step embeddings. \method\ retains sequence order, diffusion-step embeddings, and the temporal processing path. Thus, its conditioning restriction is not a removal of temporal information.

Selected embedding groups are zeroed rather than deleted. With $d_t=128$ and $d_f=16$, all variants retain $128+16+1+1=146$ side channels and the same nominal parameter allocation. This construction controls for changes in network width. The full-side model can represent the restricted mapping by ignoring its position channels; the intervention instead makes that restriction explicit during optimization.

\subsection{Temporal--Feature Denoising}
In the expanded implementation, the denoiser combines ordered sequence processing with interactions along the two axes of a multivariate window. A pointwise projection maps the two-channel diffusion input to a hidden representation of width $h=64$. Each of $L=3$ residual blocks first adds a projected diffusion-step embedding and applies a structured state-space layer \citep{gu2022s4}. In the implementation, this layer processes the feature-major serialization with index $dT+t$ and length $DT$, before the tensor is reshaped for axis-specific attention.

For the resulting hidden state $\widetilde H_\ell$, temporal attention operates across time separately for each variable, while feature attention operates across variables at each time step:
\begin{align}
 A_\ell^t&=\operatorname{Attn}_t(\widetilde H_\ell),\nonumber\\
 A_\ell^f&=\operatorname{Attn}_f(\widetilde H_\ell).
 \label{eq:axes}
\end{align}
The two representations interact through
\begin{equation}
 G_\ell=\operatorname{sigmoid}(A_\ell^t)
               \odot\tanh(A_\ell^f).
 \label{eq:axisfusion}
\end{equation}
This multiplicative combination allows the temporal representation to modulate the feature representation. The fused state and side tensor are then projected to $2h$ channels:
\begin{align}
 [Q_\ell;R_\ell]&=W_gG_\ell+W_ss^{(a,b)},\nonumber\\
 V_\ell&=\operatorname{sigmoid}(Q_\ell)\odot\tanh(R_\ell).
 \label{eq:sidefusion}
\end{align}
Bias terms are omitted for clarity. A final projection of $V_\ell$ produces residual and skip outputs. Residual updates are normalized by $\sqrt{2}$, and the sum of skip outputs by $\sqrt{L}$, before the final noise projection.

The side tensor conditions each residual block on the observed grid, while axis-specific attention provides complementary routes for information exchange. These architectural operations are shared by all side-information variants. The controlled contribution evaluated here is the conditioning policy in Eq.~\eqref{eq:side}, not a newly introduced attention or gating primitive.

\subsection{Learning and Sampling}
For a noise schedule $\beta_k$, let $\alpha_k=1-\beta_k$ and $\bar\alpha_k=\prod_{j=1}^k\alpha_j$. Training uses the standard diffusion noising process \citep{ho2020ddpm,tashiro2021csdi}:
\begin{equation}
 Z_k=\sqrt{\bar\alpha_k}X+\sqrt{1-\bar\alpha_k}\,\epsilon,
 \quad \epsilon\sim\mathcal{N}(0,I).
 \label{eq:forward}
\end{equation}
The denoiser input separates clean observed values from noisy unavailable values,
\begin{equation}
 U_k=[X^C;\,(1-C)\odot Z_k],
 \label{eq:input}
\end{equation}
and predicts noise as $\epsilon_\theta(U_k,s^{(a,b)},k)$. The objective is
\begin{equation}
 \mathcal{L}(\theta)=\mathbb{E}_{X,C,k,\epsilon}
 \!\left[\frac{\|E\odot(\epsilon-\epsilon_\theta)\|_F^2}
 {\max(1,\|E\|_1)}\right].
 \label{eq:loss}
\end{equation}
The numerator and denominator are accumulated over each minibatch. Hidden measured values enter the training noising process, but not the clean conditioning channels. The experiment uses no auxiliary reconstruction loss.

Sampling starts from Gaussian noise and applies
\begin{equation}
 Z_{k-1}=\frac{Z_k-\beta_k\epsilon_\theta/\sqrt{1-\bar\alpha_k}}
 {\sqrt{\alpha_k}}+\sigma_k\xi_k,
 \label{eq:reverse}
\end{equation}
where $\sigma_k^2=\beta_k(1-\bar\alpha_{k-1})/(1-\bar\alpha_k)$ for $k>1$, and the final step adds no noise. The visible values are supplied through Eq.~\eqref{eq:input} at every step. All variants use the same reverse process.

We draw $S=20$ samples with $K=50$ reverse steps. The point estimate is the entrywise lower median, the tenth order statistic for $S=20$. A completed window is $C\odot X+(1-C)\odot\widehat X$, preserving the observed entries. Empirical quantiles define prediction intervals whose coverage and width are evaluated separately from point error. No post-hoc calibration transformation is applied in the expanded evaluation.

\section{Experiments}
\label{sec:experiments}
We distinguish the archived block-missing experiment summarized in the abstract from an expanded common-protocol evaluation. The former establishes the source of the abstract's numerical claims; the latter examines reconstruction, conditioning ablations, and predictive distributions in greater detail. Comparisons share targets and conditioning observations \emph{within the expanded evaluation}; results from the two experiments are not pooled or compared numerically across protocols.

\subsection{Archived Block-Missing Experiment}
\label{sec:archived}
The archived experiment uses ItalyAir with 13 variables, non-overlapping length-32 windows, and nominal 50\% structured block missingness. Training uses 150 epochs, batch size 16, and 50 diffusion steps. At inference, 20 samples are generated and their median forms the point reconstruction. The reported RMSE statistics summarize seeds 1000--1002.

\begin{table}[htbp]
\centering\small
\caption{Archived ItalyAir experiment reported in the ICONIP extended abstract. RMSE mean $\pm$ SD over seeds 1000--1002; values are reproduced from that report, not recomputed from the expanded evaluation. Bold marks the lowest reported mean.}
\label{tab:archived}
\begin{tabular}{@{}lr@{}}
\toprule
Method & RMSE $\downarrow$ \\
\midrule
TCN-diff (local) & $0.834\pm0.010$ \\
DilatedConv-diff (local) & $0.693\pm0.007$ \\
CSDI (local adaptation) & $0.355\pm0.020$ \\
Full-context counterpart & $0.355\pm0.006$ \\
Block-specific feature-side & $\mathbf{0.340\pm0.020}$ \\
\bottomrule
\end{tabular}

\end{table}

Table~\ref{tab:archived} reports RMSE $0.340\pm0.020$ for block-specific feature-side conditioning, compared with $0.355\pm0.006$ for the full-context counterpart and $0.355\pm0.020$ for local CSDI. The two simpler local diffusion-denoiser controls have higher reported error. This is the conditioning comparison described in the abstract. It was selected post hoc from archived runs; the following experiment examines the same conditioning choice separately.

The archived local CSDI is distinct from the official-implementation comparator evaluated below. Equivalence of the archived and expanded preprocessing, mask realizations, and evaluation implementations has not been established. Accordingly, the archived statistics are retained as reported evidence and are not used to overwrite or augment the expanded result matrix.

\begin{table}[tbp]
\centering\small
\caption{Expanded evaluation: reconstruction on ItalyAir with 50\% block-structured artificial missingness. Neural results are mean $\pm$ sample SD over three training seeds; deterministic methods have no seed SD. All rows use the same targets. Bold marks the lowest mean in each column.}
\label{tab:main}
\begingroup\setlength{\tabcolsep}{3pt}
\begin{tabular}{@{}lcc@{}}
\toprule
Method & RMSE $\downarrow$ & MAE $\downarrow$ \\
\midrule
Mean & $1.5315$ & $1.2120$ \\
Linear interpolation & $0.7720$ & $0.4862$ \\
\midrule
CSDI (adapted) & $\mathbf{\pms{0.4127}{0.0022}}$ & $\mathbf{\pms{0.2612}{0.0018}}$ \\
\midrule
Full-side & $\pms{0.4792}{0.0187}$ & $\pms{0.3259}{0.0083}$ \\
No-side & $\pms{0.4641}{0.0111}$ & $\pms{0.3122}{0.0066}$ \\
\method\ (feature-only) & $\pms{0.4524}{0.0148}$ & $\pms{0.3131}{0.0131}$ \\
\bottomrule
\end{tabular}
\endgroup

\end{table}

\subsection{Expanded Evaluation Setup}
\label{sec:expanded-setup}
\paragraph{Dataset and preprocessing.}
ItalyAir is derived from the UCI Air Quality dataset \citep{vito2008air}. The processed sequence contains 9,357 timestamped rows and 13 measurement variables. A chronological split assigns 5,614/1,871/1,872 rows to training, validation, and testing, respectively. Non-overlapping windows of length $T=32$ yield 175/58/58 complete windows. Missingness is identified before normalization, and each variable is standardized using valid training observations only. Native missing entries are excluded from supervised targets. NMHC(GT) is retained as an input channel but has no evaluable test targets.

\paragraph{Structured masks.}
The test mask begins with 17 potentially overlapping $4\times3$ rectangles in each time--variable window. A split-level point adjustment then makes the number of hidden valid observations exactly 50\% of the observed set. Thus, the evaluation combines rectangular gaps with a point adjustment, rather than consisting exclusively of intact rectangles. The 58 test windows contain 21,292 measured cells, partitioned into 10,646 conditioning cells and 10,646 targets. The same saved masks are used for every method.

\paragraph{Compared methods.}
We compare \method\ with training-mean imputation and within-window linear interpolation. Two architecture-matched variants, Full-side and No-side, isolate the effects of positional exposure and variable identity. We also evaluate the authors' official CSDI implementation \citep{tashiro2021csdi}, adapted to the common data, fixed training masks, and evaluation procedure. CSDI retains its four-layer, width-64, eight-head nonlinear-attention denoiser; the internal variants use three layers, width 64, and 16 heads. CSDI is therefore an external comparator, not a parameter-matched ablation.

\paragraph{Training and evaluation.}
Every neural configuration in this expanded evaluation is trained with seeds 2101, 2102, and 2103 for 150 epochs, using batch size 16 and Adam \citep{kingma2015adam}. Each run performs 1,650 optimizer updates and uses its final-epoch checkpoint. The internal variants share the fixed training conditioning mask, optimization schedule, diffusion objective, and sampler. At test time, all neural methods generate 20 samples with 50 reverse steps and common window-indexed random draws. Point estimates are computed within each run before metrics are averaged across seeds. The CSDI results are same-protocol measurements, not scores from the original CSDI benchmark.

\paragraph{Metrics.}
RMSE and MAE are computed over the $N=10{,}646$ target cells in standardized units. We report the mean and sample standard deviation of the three run-level metrics. Distributional evaluation uses empirical CRPS (eCRPS) from the 20-sample empirical predictive distribution and the coverage and mean width of central 80\% and 90\% intervals. Interval endpoints use linearly interpolated empirical quantiles. All point and distributional metrics are first computed within each independently trained run and then summarized across the three training seeds.

\begin{table}[tbp]
\centering\small
\caption{Expanded evaluation: matched-seed effects of feature-side conditioning. Differences are \method\ minus the named comparator; negative values favor \method. All listed configurations share the backbone, training budget, and sampler.}
\label{tab:paired}
\begin{tabular}{@{}lrrr@{}}
\toprule
Comparator & Seed & $\Delta$RMSE & $\Delta$MAE \\
\midrule
Full-side & 2101 & $-0.0399$ & $-0.0167$ \\
 & 2102 & $-0.0165$ & $-0.0159$ \\
 & 2103 & $-0.0241$ & $-0.0057$ \\
\midrule
No-side & 2101 & $-0.0274$ & $-0.0079$ \\
 & 2102 & $-0.0106$ & $-0.0069$ \\
 & 2103 & $+0.0030$ & $+0.0177$ \\
\bottomrule
\end{tabular}

\end{table}

\begin{table*}[tbp]
\centering\small
\caption{Expanded evaluation: distributional performance from the same 20 samples per run. eCRPS and interval widths use standardized units; coverage is reported as a percentage. Values are mean $\pm$ sample SD over three seeds. Lower eCRPS is better; coverage is compared with the nominal 80\% and 90\% levels, and width is interpreted jointly with coverage.}
\label{tab:probabilistic}
\begingroup\setlength{\tabcolsep}{3pt}
\begin{tabular}{@{}lccccc@{}}
\toprule
Method & eCRPS $\downarrow$ & Coverage 80 (\%) & Width 80 & Coverage 90 (\%) & Width 90 \\
\midrule
CSDI (adapted) & $\mathbf{\pms{0.1984}{0.0018}}$ & $\pms{53.82}{1.20}$ & $\pms{0.4648}{0.0050}$ & $\pms{63.03}{0.95}$ & $\pms{0.5779}{0.0056}$ \\
Full-side & $\pms{0.2558}{0.0071}$ & $\pms{45.13}{0.83}$ & $\pms{0.4730}{0.0047}$ & $\pms{53.52}{0.95}$ & $\pms{0.5879}{0.0072}$ \\
No-side & $\pms{0.2421}{0.0057}$ & $\pms{51.10}{1.02}$ & $\pms{0.5367}{0.0109}$ & $\pms{59.83}{0.86}$ & $\pms{0.6723}{0.0134}$ \\
\method\ (feature-only) & $\pms{0.2470}{0.0120}$ & $\pms{44.81}{2.15}$ & $\pms{0.4639}{0.0044}$ & $\pms{52.89}{1.99}$ & $\pms{0.5789}{0.0074}$ \\
\bottomrule
\end{tabular}
\endgroup

\end{table*}

\subsection{Expanded Reconstruction Results}
In the expanded evaluation, Table~\ref{tab:main} shows that \method\ achieves RMSE \FeatRMSE\ and MAE \FeatMAE. Relative to linear interpolation, these correspond to reductions of \GainLinearRMSE\% and \GainLinearMAE\%, respectively. The gap indicates that the learned multivariate reconstruction is substantially more effective than local interpolation on the evaluated target set. The comparison with Full-side is more specific: without changing the backbone or training budget, feature-side conditioning reduces RMSE by \GainFullRMSE\% and MAE by \GainFullMAE\%. This result supports selective conditioning as a useful design choice within the shared diffusion model.

CSDI achieves the strongest aggregate performance, with RMSE \CSDIRMSE\ and MAE \CSDIMAE. Consequently, the internal gain from feature-side conditioning does not establish superiority over the external diffusion baseline. The comparison separates two findings: side-information exposure materially affects the shared backbone, while backbone choice remains important to the overall reconstruction quality.

\begin{table*}[tbp]
\centering\small
\caption{Expanded evaluation: per-variable RMSE, averaged over training seeds for neural methods. $N_d$ is the number of evaluated cells. All 13 input variables are retained; \NA\ denotes no evaluable target. Bold marks the lowest mean in each evaluable row.}
\label{tab:perfeature}
\begingroup\setlength{\tabcolsep}{4pt}
\begin{tabular}{@{}lrrrrrrr@{}}
\toprule
Variable & $N_d$ & Mean & Linear & CSDI & Full-side & No-side & \method \\
\midrule
CO(GT) & 575 & $1.0055$ & $0.7165$ & $0.4201$ & $0.3676$ & $\mathbf{0.3473}$ & $0.3581$ \\
PT08.S1(CO) & 810 & $0.9616$ & $0.5487$ & $\mathbf{0.2455}$ & $0.2778$ & $0.2838$ & $0.2611$ \\
NMHC(GT) & 0 & \NA & \NA & \NA & \NA & \NA & \NA \\
C6H6(GT) & 1,024 & $0.9829$ & $0.6667$ & $\mathbf{0.2454}$ & $0.2707$ & $0.2798$ & $0.2772$ \\
PT08.S2(NMHC) & 1,043 & $1.1086$ & $0.7365$ & $\mathbf{0.2563}$ & $0.2822$ & $0.2852$ & $0.2798$ \\
NOx(GT) & 976 & $2.0581$ & $1.2259$ & $0.6464$ & $0.7111$ & $0.7189$ & $\mathbf{0.6306}$ \\
PT08.S3(NOx) & 948 & $1.1150$ & $0.6970$ & $\mathbf{0.3207}$ & $0.3630$ & $0.3364$ & $0.3591$ \\
NO2(GT) & 985 & $1.9395$ & $1.1960$ & $\mathbf{0.6369}$ & $0.7952$ & $0.6866$ & $0.7068$ \\
PT08.S4(NO2) & 986 & $2.0998$ & $0.6446$ & $\mathbf{0.3287}$ & $0.6228$ & $0.6067$ & $0.6100$ \\
PT08.S5(O3) & 1,005 & $1.2153$ & $0.8553$ & $0.4097$ & $0.3928$ & $0.3906$ & $\mathbf{0.3821}$ \\
T & 985 & $1.9619$ & $0.4059$ & $\mathbf{0.3948}$ & $0.4104$ & $0.4318$ & $0.4295$ \\
RH & 769 & $1.0402$ & $\mathbf{0.4179}$ & $0.4899$ & $0.4518$ & $0.4584$ & $0.4401$ \\
AH & 540 & $1.8738$ & $\mathbf{0.1138}$ & $0.1552$ & $0.2676$ & $0.2937$ & $0.2749$ \\
\bottomrule
\end{tabular}
\endgroup

\end{table*}

\subsection{Side-Information Ablation}
Table~\ref{tab:paired} resolves the average improvement into matched training runs. Relative to Full-side, \method\ improves both RMSE and MAE for all three seeds. RMSE differences range from $-0.0399$ to $-0.0165$, so the average improvement is not driven by a single favorable initialization. The finding is consistent with the proposed distinction between temporal processing and auxiliary positional exposure: the temporal operators remain useful even when the explicit position group is suppressed.

The No-side comparison isolates variable identity after position embeddings have already been removed. \method\ has \GainNoRMSE\% lower mean RMSE, but its mean MAE is slightly higher (\FeatMAE\ versus \NoMAE). It improves both metrics in two seeds and worsens both in seed 2103. These results indicate that retaining variable identity changes the distribution of reconstruction errors rather than uniformly improving every metric. Because RMSE weights larger errors more strongly than MAE, the aggregate ordering is consistent with an advantage concentrated in larger deviations; variable-level results provide a more localized view.

\begin{figure*}[tbp]
\centering
\includegraphics[width=.96\textwidth]{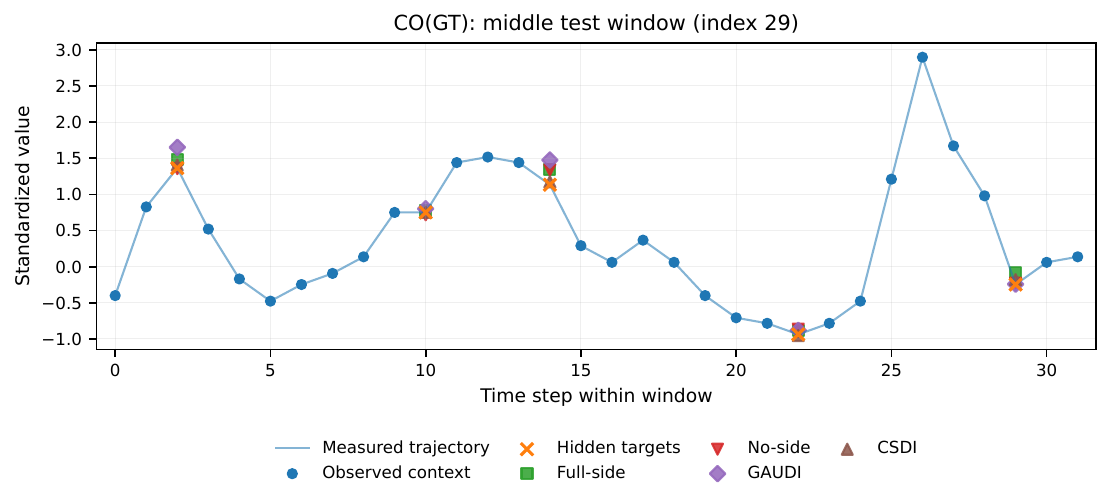}
\caption{Expanded evaluation: CO(GT) reconstruction in the chronologically selected middle test window (ordinal 29). Curves and markers use the seed-2101 models, not an ensemble across training seeds. Predictions are shown only at artificially hidden targets; observed values remain available as context. The displayed window is selected chronologically rather than by relative method performance.}
\label{fig:case}
\end{figure*}

\subsection{Probabilistic Performance and Calibration}
Table~\ref{tab:probabilistic} shows that \method\ lowers eCRPS from 0.2558 to 0.2470 relative to Full-side, accompanying the improvement in point reconstruction. No-side achieves a lower eCRPS of 0.2421 despite its higher RMSE. Thus, the conditioning policy preferred by point reconstruction need not be the one preferred by a distributional score. CSDI again provides the strongest aggregate eCRPS, at 0.1984.

Coverage reveals a different issue. \method's nominal 90\% interval covers only 52.89\% of targets, with mean width 0.5789. The corresponding coverages are 53.52\% for Full-side, 59.83\% for No-side, and 63.03\% for CSDI. All four methods substantially under-cover. The wider No-side intervals improve coverage but remain far from the nominal level, while the narrower feature-side intervals do not constitute a calibration gain. These measurements show that selective conditioning improves reconstruction relative to Full-side without resolving predictive undercoverage. Reliable intervals require additional calibration or distributional modeling beyond the tested conditioning change.

\subsection{Variable-Level Behavior}
Table~\ref{tab:perfeature} identifies where the aggregate gain arises. \method\ improves over Full-side on nine of the twelve evaluable variables. The largest absolute reductions occur for NO2(GT), from 0.7952 to 0.7068, and NOx(GT), from 0.7111 to 0.6306. These channels illustrate the benefit of changing the conditioning policy while retaining joint temporal--feature processing. The effect is not uniform: C6H6(GT), temperature, and absolute humidity have higher RMSE under feature-side conditioning.

The external comparison is also heterogeneous. \method\ improves over CSDI on CO(GT), NOx(GT), PT08.S5(O3), and relative humidity, whereas CSDI remains stronger on most other channels and in aggregate. Linear interpolation performs best on absolute humidity, showing that a simple temporal model can remain competitive for an individual variable even when its aggregate error is much higher. This pattern favors interpreting the contribution as a conditioning improvement with channel-dependent effects, rather than a uniform advantage across measurement types.

\subsection{Qualitative Reconstruction}
Figure~\ref{fig:case} complements the aggregate tables with a fixed chronological example. The observed trajectory provides context for the hidden values, while the target-only prediction markers expose local differences without treating native missing entries as measured ground truth. The example includes both close reconstructions and deviations around hidden changes in the signal. Such local behavior explains why a favorable aggregate ordering should be read together with variable-level and distributional results. The displayed case is the chronologically selected middle test window rather than a window ranked by which method performs best.

\FloatBarrier
\section{Conclusion}
We studied a block-specific, \method-aligned conditional diffusion imputer that separates observation geometry and variable identity from auxiliary within-window coordinates. Feature-side conditioning preserves state-space and temporal--feature modeling while restricting the side information exposed to the denoiser. The archived experiment supplies the block-missing result reported in the abstract. In the separate expanded ItalyAir evaluation, this choice reduces mean RMSE by \GainFullRMSE\% relative to full-side conditioning, with improvements in all three matched seeds. The variable-level and probabilistic analyses further show that conditioning choices affect error types and interval behavior differently. The central finding is that selective side-information exposure can improve diffusion reconstruction without changing the backbone, but point accuracy and uncertainty calibration remain distinct objectives.

\paragraph{Future work.}
The evaluation covers one dataset, one window length, and one block-structured mask suite. Broader tests should vary datasets and missingness geometries, include the remaining time-only conditioning control, and examine calibration using a held-out calibration procedure. These extensions would establish the transferability of the conditioning effect and address the undercoverage observed here.

\begingroup\footnotesize
\setlength{\bibsep}{1.5pt}
\bibliographystyle{abbrvnat}
\bibliography{references}
\endgroup
\balance
\end{document}